\documentclass[letterpaper, 10 pt, conference]{ieeeconf}  

\usepackage{graphicx}
\usepackage{algpseudocode}  
\usepackage{booktabs} 
\usepackage{multirow} 
\usepackage{amsmath}  
\usepackage{tcolorbox} 
\usepackage{amsfonts} 
\usepackage{makecell} 
\usepackage{pifont}  
\usepackage{bm}  
\usepackage{todonotes} 
\usepackage{stfloats}  
\usepackage{algorithm}
\usepackage{algpseudocode}
\usepackage{cite}

\IEEEoverridecommandlockouts                              

\title{\LARGE \bf
Whole-Body Inverse Kinematics with Graph Diffusion
}

\author{Helong Huang$^{1}$,
Kai Tan$^{1}$,
Feng Wen$^{1}$,
Guowei Huang$^{1}$,
Xingyue Quan$^{1}$
\thanks{$^{1}$Large Model Algorithm Lab, Huawei}%
}
\begin{document}

\maketitle
\thispagestyle{empty}
\pagestyle{empty}


\begin{abstract}

Inverse kinematics (IK) is a fundamental problem in robotics, requiring the generation of joint configurations that satisfy target end-effector poses. Existing approaches often struggle to generalize across diverse robot morphologies and to effectively model the multi-modal nature of IK, particularly in articulated systems with multiple kinematic branches. In this work, we propose GraphDiff-IK, a structure-aware graph diffusion framework for inverse kinematics. Specifically, we represent the robot as a kinematic graph constructed from the robot URDF, where nodes correspond to actuated joints and edges encode kinematic dependencies. Building upon this representation, we formulate IK as a conditional graph diffusion process that directly generates joint configurations on the robot graph. To better capture structural dependencies in articulated systems, we further introduce a structure-aware graph reasoning framework with hierarchical stage-wise message passing and torso-aware conditioning for multi-branch robots. In addition, we incorporate noisy forward kinematics feedback and task-space supervision to improve geometric consistency during denoising. The proposed framework provides a unified formulation that naturally supports single-arm robots, dual-arm systems, and articulated robots with torso or waist structures. Extensive experiments on diverse robotic platforms demonstrate that the proposed method achieves accurate and stable IK performance while preserving the ability to generate multiple feasible solutions for redundant robotic systems.

\end{abstract}

\begin{keywords}
inverse kinematics, diffusion model, graph neural network, articulated robots, multi-branch robotic systems
\end{keywords}

\section{INTRODUCTION}

\begin{figure*}[!t]
  \centering
  \includegraphics[width=\textwidth]{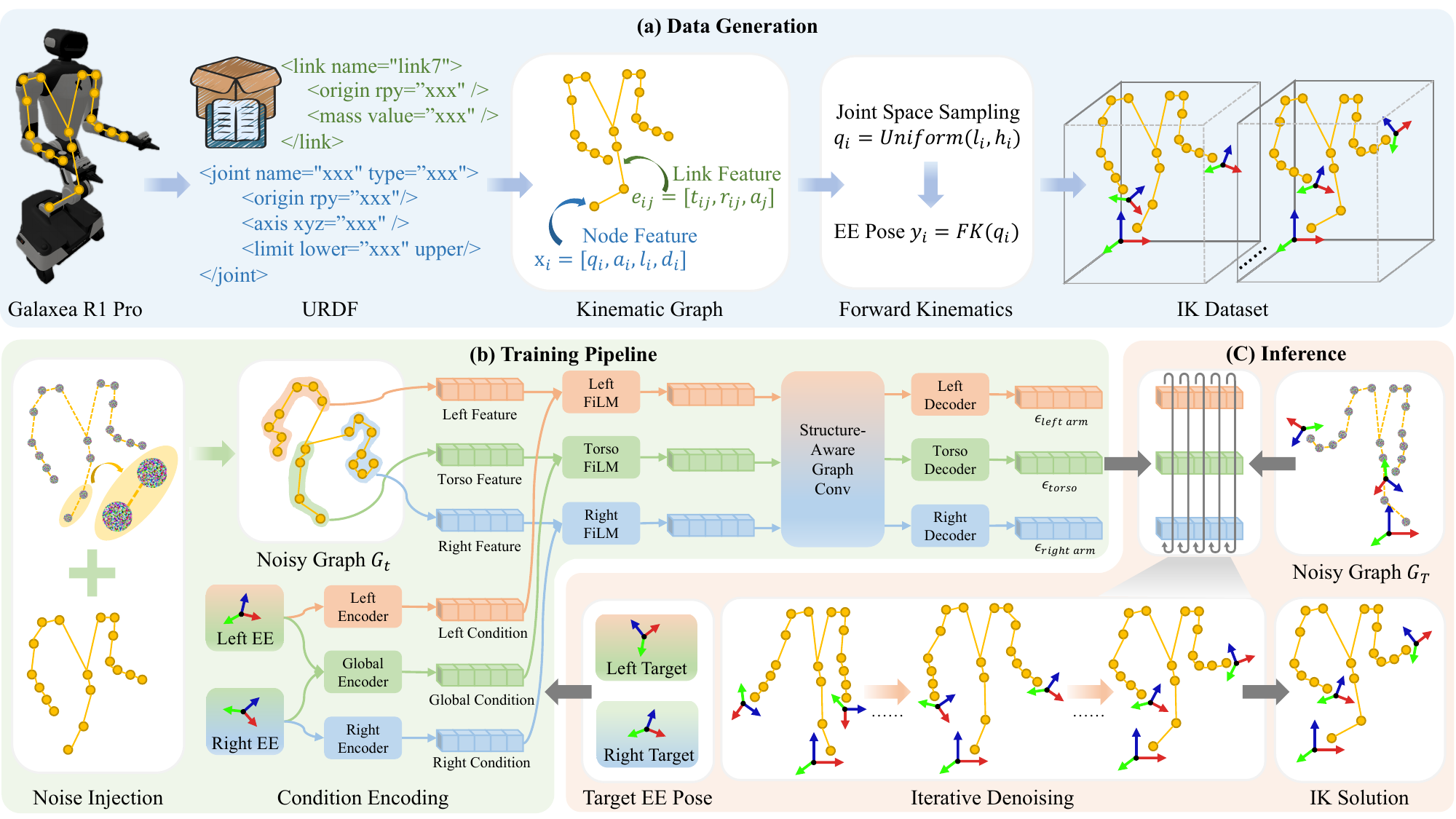}
  \caption{\textbf{Overview of GraphDiff-IK.}
  (a) \textbf{Data generation.} The robot URDF is converted into a kinematic graph, where nodes represent actuated joints and edges encode kinematic relations. Joint configurations are sampled from the joint limits, and corresponding end-effector poses are computed via forward kinematics to construct an IK dataset of pairs $(q, y)$.
  (b) \textbf{Training pipeline.} Given a clean graph $G_0$, forward diffusion progressively corrupts joint states to obtain a noisy graph $G_t$. The model takes the noisy graph together with condition encodings derived from the target end-effector poses, and predicts structured noise $\epsilon = [\epsilon_{\text{left arm}}, \epsilon_{\text{torso}}, \epsilon_{\text{right arm}}]$ using a branch-aware graph diffusion network with FiLM conditioning. The model is trained to reverse the diffusion process by denoising the graph while respecting the robot kinematic structure.
  (c) \textbf{Inference.} Starting from an initial Gaussian noise graph $G_T$, the model iteratively performs reverse diffusion conditioned on the target end-effector pose to generate a valid joint configuration $\hat{q}$, enabling inverse kinematics without explicit optimization.
}
  \label{fig:overview}
\end{figure*}




Inverse kinematics (IK) is a fundamental problem in robotic control, aiming to compute joint configurations that achieve a desired end-effector pose~\cite{siciliano2009robotics, merat1987introduction, spong2020robot}. It plays a critical role in a wide range of applications, including manipulation, motion planning, and human–robot interaction~\cite{lavalle2006planning}, serving as a key bridge between high-level task specifications and low-level control.

With the rapid evolution of robotic platforms, systems are transitioning from traditional single-arm manipulators~\cite{spong2020robot} to more complex configurations with significantly higher degrees of freedom, such as dual-arm robots~\cite{lowcost_bimanual_2023}, humanoid robots with articulated torsos~\cite{unitree_g1}, and even whole-body systems integrated with mobile bases~\cite{li2024behavior1k}. While these systems offer enhanced expressiveness and capability, they also substantially increase the complexity of the IK problem. High degrees of freedom introduce redundancy, multi-branch kinematic structures lead to intricate coupling effects, and coordinating multiple end-effectors further complicates the solution space, making IK increasingly challenging to solve in a stable and efficient manner. As a result, traditional IK methods face significant challenges when applied to such complex robotic systems~\cite{aristidou2018inverse}.






Existing approaches to IK can be broadly categorized into classical methods based on analytical or numerical optimization, and data-driven learning-based methods~\cite{siciliano2009robotics, wampler1986manipulator}. Classical IK methods rely on accurate kinematic models and typically solve the problem using Jacobian-based updates, optimization techniques, or closed-form derivations. While effective for simple kinematic chains, these methods often struggle in high-dimensional or multi-branch systems, where they can become unstable, sensitive to initialization, and difficult to extend to complex constraints.

More recently, learning-based approaches have been proposed to directly learn mappings from end-effector poses to joint configurations using neural networks, offering improved inference efficiency~\cite{ggik, ikflow, iknet}. However, most existing methods are designed for fixed robot structures, predominantly focusing on single-arm systems, and lack explicit modeling of kinematic topology. As a result, they often fail to generalize to multi-arm systems or robots with torso or waist articulation. In addition, these methods typically struggle to balance accuracy and stability, with noticeable errors in both end-effector position and orientation, limiting their applicability in high-precision manipulation tasks.

Overall, existing methods lack a unified way to model complex, multi-branch kinematic structures and have limited ability to generalize across different robot morphologies, which remains a key bottleneck for applying IK in modern high-degree-of-freedom robotic systems. As summarized in Table~\ref{tab:capability}, most existing approaches satisfy only a subset of these requirements, and none simultaneously achieve structure-aware modeling, multi-end-effector control, and multi-solution IK modeling.



 
To address these challenges, we reformulate inverse kinematics as a structured generative modeling problem and propose a structure-aware graph diffusion framework for inverse kinematics, termed \textbf{GraphDiff-IK}. The key idea is to represent the robot as a kinematic graph and model inverse kinematics as a structured generative process over this graph, allowing the model to explicitly preserve kinematic relationships during generation.

Specifically, we convert the robot URDF into a kinematic graph, where nodes correspond to joints and edges encode their kinematic dependencies. A diffusion model is then applied directly in the joint space~\cite{ddpm}, progressively denoising sampled states to generate joint configurations that satisfy the desired end-effector constraints. To better exploit structural information, we further introduce a structure-aware graph convolution mechanism, together with a staged modeling strategy that captures dependencies across different components, such as the torso and the arms. This enables effective modeling of coupling in complex multi-branch systems.



The proposed approach offers several advantages. First, by formulating IK as a generative process, it avoids explicit reliance on analytical formulations or iterative optimization procedures, leading to improved robustness in complex, high-dimensional systems. Second, diffusion models naturally capture the multi-solution nature of IK~\cite{chi2023diffusionpolicy}, enabling the generation of multiple valid joint configurations that satisfy the same end-effector constraints, thereby improving solution diversity and robustness. Finally, the graph-based representation allows the model to explicitly preserve kinematic structure throughout the generation process, making it adaptable to different robot morphologies and capable of generalizing across diverse systems. The main contributions of this work are summarized as follows:

\begin{itemize}
    
    \item We propose a unified graph-based formulation for inverse kinematics, where robots with diverse kinematic topologies, such as single-arm, dual-arm, and torso- or waist-equipped robots, are modeled as structured graphs, enabling a consistent formulation across different robot morphologies.
    
    \item We introduce a structure-aware graph diffusion model that explicitly preserves kinematic structure during the generative process, improving modeling capability for multi-branch and high-degree-of-freedom systems.
    
    \item We design a staged structure modeling mechanism, where torso-aware conditioning is leveraged to guide multi-arm generation, effectively capturing coupling among different structural components.

    \item We formulate inverse kinematics as a generative modeling problem, allowing the model to capture the distribution over feasible joint configurations and generate multiple diverse and valid solutions under end-effector constraints.

    \item Extensive experiments on a variety of robotic platforms demonstrate that the proposed method outperforms existing approaches in terms of end-effector position and orientation accuracy, as well as cross-morphology generalization.
\end{itemize}

\begin{table}[t]
\caption{Comparison of inverse kinematics methods in terms of structural modeling and capability.}
\label{tab:capability}
\centering
\setlength{\tabcolsep}{4pt}
\renewcommand{\arraystretch}{1.2}
\begin{tabular}{l c c c}
\toprule
\textbf{Method} 
& \shortstack{\textbf{Structure-Aware} \\ \textbf{Modeling}} 
& \shortstack{\textbf{Multi-EE} \\ \textbf{Control}} 
& \shortstack{\textbf{Multi-Solution} \\ \textbf{IK Modeling}} \\
\midrule
Transformer~\cite{transformer}  
& $\times$ & $\checkmark$ & $\times$ \\
GNN~\cite{kipf2016semi}
& $\checkmark$ & $\checkmark$ & $\times$ \\
MLP~\cite{rumelhart1986learning}
& $\times$ & $\checkmark$ & $\times$ \\
IKNet~\cite{iknet}      
& $\times$ & $\times$ & $\checkmark$ \\
IKFlow~\cite{ikflow}     
& $\times$ & $\times$ & $\checkmark$ \\
GGIK~\cite{ggik}       
& $\checkmark$ & $\times$ & $\checkmark$ \\
\midrule
\textbf{GraphDiff-IK (Ours)} 
& $\checkmark$ 
& $\checkmark$ 
& $\checkmark$ \\
\bottomrule
\end{tabular}
\end{table}


\section{Related Work}

\subsection{Classical Inverse Kinematics}

Classical inverse kinematics methods primarily solve joint configurations through analytical derivation or numerical optimization~\cite{siciliano2009robotics, craig2005introduction, spong2020robot}. Analytical approaches derive closed-form solutions from robot kinematic equations and can achieve efficient and accurate IK solving under specific kinematic structures~\cite{pieper1969kinematics}. However, such methods are highly dependent on manually derived formulations and are generally limited to robots with relatively regular structures or low degrees of freedom. In contrast, numerical approaches typically formulate IK as an optimization problem and iteratively approximate the target pose through Jacobian-based methods, pseudo-inverse solvers, or gradient-based optimization techniques~\cite{wampler1986manipulator, nakamura1986inverse, buss2004introduction}.

Although classical methods have demonstrated stable performance on low-dimensional or structurally simple robotic systems, their performance often becomes limited when robots exhibit high degrees of freedom, kinematic redundancy, or multi-branch kinematic chains~\cite{yoshikawa1985manipulability, chiaverini2002singularity}. On the one hand, iterative optimization procedures are usually sensitive to initialization and can suffer from local minima or kinematic singularities~\cite{wampler1986manipulator, chiaverini2002singularity}. On the other hand, these methods strongly rely on accurate kinematic models and often exhibit limited robustness under modeling inaccuracies or complex articulated structures. More importantly, for robotic systems with shared torso, waist or dual-arm coordination structures, strong structural coupling exists among different kinematic branches, further increasing the complexity of optimization~\cite{sentis2005control}. As robotic systems continue to evolve to high-dimensional, topologically complex, and multi-branch morphologies, classical IK methods become increasingly difficult to generalize within a unified framework~\cite{siciliano2009robotics, khatib1987unified}.

\subsection{Learning-based Inverse Kinematics}

To overcome the limitations of iterative optimization in computational efficiency and complex structure modeling, recent studies have introduced deep learning approaches to directly learn mappings from end-effector poses to joint configurations~\cite{csiszar2017solving, ho2023deep}. These methods typically formulate IK as a supervised learning problem and employ deep neural networks, such as multilayer perceptrons (MLPs)~\cite{rumelhart1986learning} and Transformers~\cite{transformer}, to directly predict target joint configurations through feed-forward inference. Compared with conventional numerical optimization methods, learning-based approaches replace online iterative solving with offline training, thereby significantly improving inference efficiency. However, since inverse kinematics inherently exhibits one-to-many mapping characteristics, deterministic regression-based approaches often struggle to effectively model diverse valid solutions corresponding to the same target pose, thereby limiting their ability to represent complex solution distributions~\cite{ikflow, iknet}.

To further capture the inherent multi-solution property of IK, several studies have introduced probabilistic modeling or generative learning frameworks. For example, IKNet~\cite{iknet} generates diverse IK solutions through latent-space modeling, while IKFlow~\cite{ikflow} models conditional joint distributions using neural networks, enabling the sampling of multiple valid configurations for the same target pose. These studies suggest that inverse kinematics is inherently not a strictly one-to-one regression problem, but rather a conditional generation problem with multimodal solution distributions.

Despite their promising performance, existing learning-based IK methods still exhibit several important limitations. First, most methods are designed for specific robot structures and lack unified generalization across different robot morphologies. Second, existing approaches commonly represent robot states using flattened joint vectors, ignoring the explicit topological dependencies inherently encoded in articulated robotic systems. When robots involve dual-arm structures, shared torso coordination, or multiple coupled kinematic chains, such flattened representations become insufficient for modeling long-range dependencies and structural coupling among different kinematic branches. Furthermore, most existing learning-based IK methods mainly focus on single-arm systems and still lack effective mechanisms for modeling structured coordination in high-dimensional multi-branch robotic systems.

\subsection{Graph-based Robot Modeling}

Robotic systems are inherently articulated systems with explicit topological structures, where joints and links naturally form hierarchical kinematic dependencies. Recently, graph neural networks (GNNs)~\cite{zhou2020graph, kipf2016semi} and graph-based modeling approaches have been widely adopted in robotics for system modeling and control tasks~\cite{wang2018nervenet, sanchez2018graph}. By representing joints as graph nodes and kinematic relations as graph edges, graph representations provide a natural mechanism for modeling topological structures, hierarchical dependencies, and long-range interactions among different robot components~\cite{rofdik}.

Existing graph-based methods have been successfully applied to dynamics modeling, control policy learning, multibody system modeling, and physical interaction reasoning~\cite{rofdik, ggik, vosylius2024instant}. These studies demonstrate that graph representations can effectively improve structural modeling capability in high-dimensional robotic systems. Moreover, graph representations provide a unified structured representation across different robot morphologies, thereby improving scalability and generalization across diverse kinematic topologies.

However, existing graph-based robotics methods mainly focus on forward modeling problems, such as predicting system dynamics or learning control policies from joint states~\cite{wang2018nervenet, sanchez2018graph}, while inverse kinematics remains relatively underexplored. Unlike forward prediction tasks, inverse kinematics is fundamentally a conditional generation problem, where the objective is to generate valid joint configurations conditioned on target end-effector constraints. Particularly in redundant or multi-branch robotic systems, a single target pose may correspond to multiple valid solutions, making IK inherently a one-to-many generation problem. Existing graph-based methods still predominantly adopt deterministic regression formulations and lack explicit modeling of the multimodal generative nature of IK. Therefore, how to integrate structured graph representations with generative modeling for inverse kinematics in articulated robotic systems remains largely unexplored.

\subsection{Diffusion Models for Robotics}

Diffusion models~\cite{ddpm, song2020score} have recently emerged as a powerful class of generative models for learning complex high-dimensional distributions. By progressively denoising random noise into structured samples, diffusion models have achieved remarkable success in image generation, video synthesis, trajectory generation, and decision-making tasks~\cite{ddpm, song2020score, chi2023diffusionpolicy}. In robotics, diffusion-based approaches have been widely applied to policy learning, action generation, trajectory planning, and robot control~\cite{janner2022planning, chi2023diffusionpolicy}. For example, Diffusion Policy utilizes diffusion processes to model multimodal action distributions and generate high-quality robot control policies~\cite{chi2023diffusionpolicy, wolf2025diffusion, carvalho2023motion, carvalho2025motion}.

The strong generative capability of diffusion models makes them particularly suitable for inverse problems with multimodal solution spaces. Since inverse kinematics is inherently a one-to-many and multimodal problem, where a single end-effector target may correspond to multiple valid joint configurations, diffusion models provide a natural conditional generation framework for modeling diverse IK solution distributions.

Despite their strong generative capability, existing diffusion-based robotics methods typically adopt flattened state or action representations, where robot configurations are directly represented as vectors without explicit structural awareness. Such designs become increasingly limited when applied to robotic systems with complex topological structures, hierarchical dependencies, and multi-branch kinematic chains. For example, in dual-arm robots, torso-coordinated manipulators, and humanoid robotic systems~\cite{unitree_g1}, significant structural coupling exists among different kinematic branches, while flattened representations struggle to explicitly capture these structured dependencies. Consequently, existing diffusion-based robotics methods still lack a unified structure-aware generative framework for inverse kinematics across diverse robot morphologies.

In summary, existing inverse kinematics methods still face significant limitations when dealing with high-dimensional, multi-branch, and complex articulated robotic systems. Classical optimization-based methods struggle to scale to robotic systems with strong structural coupling. Existing learning-based approaches generally lack explicit topology-aware modeling and are primarily designed for specific robot morphologies. Existing graph-based robotics methods mainly focus on forward modeling problems and rarely address the generative nature of inverse kinematics. Meanwhile, existing diffusion-based robotics methods predominantly rely on flattened representations and therefore fail to effectively capture structured dependencies in complex robotic systems. Consequently, developing a unified structure-aware generative framework for inverse kinematics across diverse robot morphologies remains a challenging open problem.

\section{PRELIMINARIES}

\subsection{Robot Kinematic Graph}

Robotic systems can be naturally represented as articulated kinematic graphs with explicit topological structures~\cite{featherstone2008rigid}. Given a robot, we represent its kinematic structure as a graph
\begin{equation}
G = (V, E),
\end{equation}
where each node $v_i \in V$ corresponds to an actuated joint and each edge $(i,j) \in E$ represents the articulated kinematic dependency between adjacent joints.

In this formulation, node features describe joint-related states and properties, such as joint configurations and joint attributes, while edge features encode relative kinematic transformations and structural relationships between connected joints. Such a graph representation naturally preserves the hierarchical topology and long-range structural dependencies within articulated robotic systems.

Compared with flattened vector representations, graph-based representations provide a more structured formulation for modeling robots with complex morphologies, including dual-arm systems, shared torso structures, and multi-branch kinematic chains. This structured representation further provides a unified formulation across different robot topologies and morphologies. For a robot with $N$ actuated joints, the joint configuration is represented as
\begin{equation}
q = [q_1, q_2, \dots, q_N] \in \mathbb{R}^N,
\end{equation}
where $q_i$ denotes the state of the $i$-th joint.

\subsection{Inverse Kinematics as Conditional Distribution Modeling}

Let $q \in \mathbb{R}^N$ denote the robot joint configuration and let $y \in SE(3)$ denote the target end-effector (EE) pose, consisting of both position and orientation components. The inverse kinematics problem aims to recover valid joint configurations conditioned on the target end-effector pose. The forward kinematics (FK) function maps joint configurations to the end-effector pose in $SE(3)$:
\begin{equation}
f: \mathbb{R}^N \rightarrow SE(3), \quad y=f(q).
\end{equation}

However, due to kinematic redundancy and structural coupling in articulated robotic systems, multiple joint configurations may correspond to the same target end-effector pose. Therefore, inverse kinematics is inherently a one-to-many mapping problem with multimodal solution distributions.

To explicitly model this multi-modality, we formulate inverse kinematics as a conditional distribution:
\begin{equation}
p(q \mid y),
\end{equation}
which models the distribution of valid joint configurations conditioned on the target end-effector pose. This probabilistic formulation provides a natural framework for representing diverse feasible IK solutions and establishes the foundation for conditional generative modeling over articulated robot graphs.

\subsection{Diffusion Models}
Diffusion models are a class of generative models that learn complex data distributions through iterative denoising processes~\cite{ddpm}. Given a clean data sample $q_0$, the forward diffusion process progressively perturbs the data by adding Gaussian noise over multiple timesteps:
\begin{equation}
q_t = \sqrt{\bar{\alpha}_t} q_0 +
\sqrt{1-\bar{\alpha}_t}\epsilon,
\quad
\epsilon \sim \mathcal{N}(0, I),
\label{eq:diffusion_forward}
\end{equation}
where $\bar{\alpha}_t$ denotes the predefined noise schedule. The reverse process is parameterized by a neural network $\epsilon_\theta(\cdot)$ that predicts the injected noise and progressively reconstructs the original data distribution through iterative denoising.

For conditional generation tasks, additional conditions can be incorporated into the denoising process to model conditional distributions. In the context of inverse kinematics, the diffusion process is conditioned on the target end-effector pose $y$, enabling the model to generate joint configurations according to the conditional distribution $p(q \mid y)$. Due to their strong capability in modeling high-dimensional multimodal distributions, diffusion models provide an effective framework for generating diverse valid inverse kinematics solutions in complex articulated robotic systems.

\section{METHODOLOGY}

\subsection{Framework Overview}

We propose GraphDiff-IK, a structure-aware graph diffusion framework for inverse kinematics that models the conditional distribution of joint configurations given target end-effector poses. The proposed framework provides a unified formulation capable of handling both single-arm and multi-branch robotic systems, such as dual-arm robots with torso structures. As illustrated in Fig.~\ref{fig:overview}, the overall pipeline consists of three main components: data generation, diffusion-based training, and inference.

\subsubsection{Data Generation}
We construct a large-scale inverse kinematics dataset using forward kinematics. Specifically, joint configurations are sampled within the joint limits defined by the robot URDF, and the corresponding end-effector poses are computed to form data pairs $(q, y)$. In parallel, the URDF is converted into a kinematic graph representation, where nodes correspond to actuated joints and edges encode kinematic relationships. This process enables efficient and fully automatic data generation without manual annotation.

\subsubsection{Diffusion-Based Training}
Given the constructed dataset, we formulate inverse kinematics as a conditional generation problem and learn the distribution $p(q \mid y)$ using a graph diffusion model. During training, joint configurations are progressively perturbed through a forward diffusion process, and the model is trained to predict the injected noise conditioned on the target end-effector pose and structural information. To effectively capture kinematic dependencies, we design a structure-aware graph convolution architecture with staged message passing, which explicitly models the hierarchical relationships among the torso and different kinematic branches. Furthermore, a forward kinematics-based task-space supervision is introduced to guide the model toward geometrically consistent solutions.

\subsubsection{Inference}
At inference time, the model starts from a Gaussian noise sample and iteratively performs reverse diffusion under the guidance of the target end-effector pose. The final output is a joint configuration that satisfies the desired task constraints. This enables solving inverse kinematics without explicit optimization, while providing efficient and scalable deployment for real robotic systems.

\subsection{FK-based IK Data Generation}

To train the noise prediction model, we construct a large-scale inverse kinematics dataset using forward kinematics (FK). The dataset consists of pairs of joint configurations and corresponding end-effector poses:
\begin{equation}
\mathcal{D} = \{(q_i, y_i)\}_{i=1}^N,
\end{equation}
where $q_i \in \mathbb{R}^D$ denotes the robot joint configuration with $D$ actuated joints, and $y_i$ represents the corresponding end-effector pose.

\subsubsection{Joint Space Sampling}
We generate joint configurations by independently sampling each actuated joint within its feasible range defined by the robot URDF:
\begin{equation}
q_i \sim \mathcal{U}(q_{\min}, q_{\max}),
\end{equation}
where $q_{\min}$ and $q_{\max}$ denote the lower and upper joint limits, respectively. This sampling strategy enables efficient coverage of the feasible joint space without requiring expert demonstrations or task-specific motion data.

\subsubsection{Forward Kinematics Mapping}
For each sampled joint configuration $q_i$, the corresponding end-effector pose is computed using forward kinematics:
\begin{equation}
y_i = \mathrm{FK}(q_i).
\end{equation}
The generated dataset therefore provides samples from the joint-to-task-space mapping underlying the conditional inverse kinematics distribution $p(q \mid y)$.

\subsubsection{Multi-Branch Systems}
For robots with multiple kinematic branches, such as dual-arm systems with torso articulation, forward kinematics is computed independently for each end-effector. The resulting target condition is therefore represented as
\begin{equation}
y_i = \left\{ \left(p_i^{(k)}, r_i^{(k)}\right) \right\}_{k=1}^K,
\end{equation}
where $K$ denotes the number of end-effectors, and $p_i^{(k)}$ and $r_i^{(k)}$ represent the position and orientation of the $k$-th end-effector, respectively. This formulation naturally captures the coordination constraints and structural dependencies across different kinematic branches.

\subsubsection{Workspace Coverage}
\begin{figure}[!t]
    \centering
    \includegraphics[width=\linewidth]{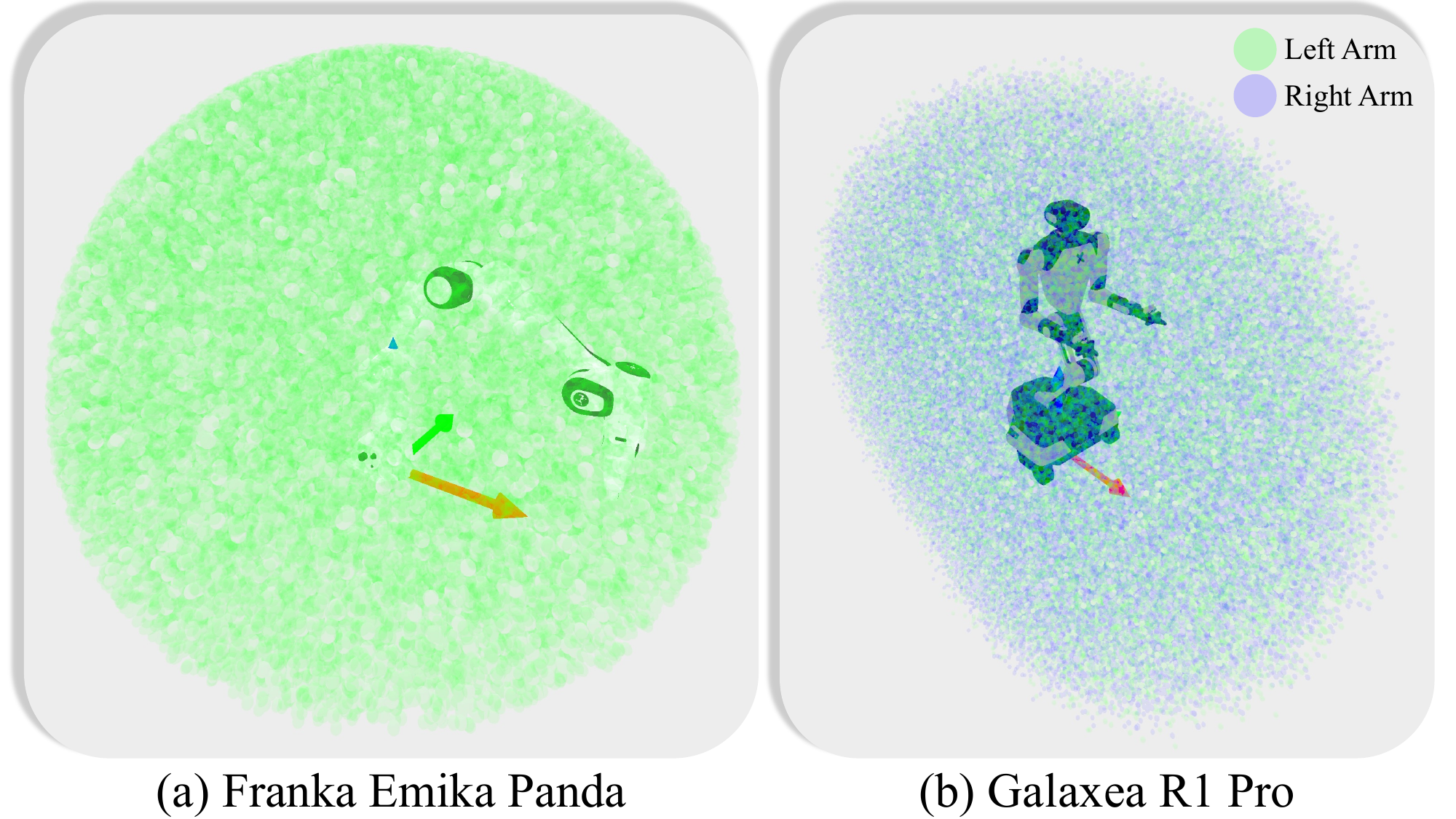}
    \caption{\textbf{End-effector workspace visualization.}
    End-effector positions generated from $10^6$ joint configurations randomly sampled within joint limits. 
    (a) Franka Emika Panda exhibits a continuous reachable workspace under single-arm kinematics. 
    (b) Galaxea R1 Pro shows a more complex distribution due to dual-arm coordination, where left and right end-effectors jointly span the workspace. The large spatial coverage highlights the high-dimensional and multi-modal nature of the inverse kinematics problem.}
    \label{fig:ee_workspace}
\end{figure}

As illustrated in Fig.~\ref{fig:ee_workspace}, random joint sampling leads to a wide spatial distribution of end-effector positions. Even for single-arm robots, the reachable workspace spans a large continuous region, while multi-branch systems exhibit more complex distributions due to inter-branch coupling and redundancy. Moreover, inverse kinematics is inherently multi-modal, where multiple distinct joint configurations may correspond to similar end-effector poses:
\begin{equation}
\exists \ q^{(a)} \neq q^{(b)}
\quad \text{s.t.} \quad
\mathrm{FK}(q^{(a)}) \approx \mathrm{FK}(q^{(b)}).
\end{equation}
This ambiguity becomes increasingly significant for redundant and multi-branch robotic systems, motivating the use of generative models for inverse kinematics.

\subsubsection{Efficiency}
The data generation process is fully analytical and does not require manual labeling or real-world data collection. In practice, the analytical formulation enables the efficient generation of large-scale datasets containing up to $10^6$ samples within a few minutes, facilitating stable training of high-capacity generative models.


\subsection{Graph Representation of Robot Kinematics}

Given a robot URDF, we represent the robot kinematic structure as a graph
\begin{equation}
G = (V, E),
\end{equation}
where each node $v_i \in V$ corresponds to an actuated joint and each edge $(i,j) \in E$ represents the kinematic dependency between two connected joints.

Compared with flattened joint representations, the proposed graph formulation explicitly preserves robot topology, local kinematic transformations, and inter-joint dependencies, enabling structured message passing and structure-aware reasoning across diverse robot morphologies, including both single-arm and multi-branch articulated robots.

\subsubsection{Node Representation}

For each node $v_i$, we define the node feature as
\begin{equation}
x_i = [q_i \, \| \, a_i \, \| \, l_i \, \| \, d_i] \in \mathbb{R}^{d_x},
\end{equation}
where $q_i$ denotes the joint angle, $a_i$ represents the joint axis, $l_i$ corresponds to the joint limits, and $d_i$ encodes structural information associated with the joint.

The structural encoding consists of two components. The first is the joint depth embedding, which captures the hierarchical position of the joint within the kinematic structure. The second is a branch embedding, which distinguishes different kinematic branches. For example, in robots with torso and dual-arm structures, torso joints and left or right arm joints are assigned different branch identifiers. For single-arm robots, only the depth embedding is required.

For multi-branch robotic systems, an additional node-type encoding is introduced to distinguish torso, left-arm, and right-arm joints. This encoding enables the construction of branch-specific subgraphs for staged message passing and hierarchical structure-aware reasoning. This design enables the diffusion model to perform structure-aware message passing and branch-specific reasoning, which is particularly important for articulated systems with hierarchical and multi-branch kinematic structures.

\subsubsection{Edge Representation}

Edges represent the kinematic relationships between joints. For each edge connecting a parent joint to a child joint, we define the edge feature as
\begin{equation}
e_{ij} = [t_{ij} \, \| \, r_{ij} \, \| \, a_j] \in \mathbb{R}^{d_e},
\end{equation}
where $t_{ij}$ and $r_{ij}$ denote the relative translation and rotation of the child joint with respect to the parent joint, and $a_j$ represents the joint axis of the child joint. These edge attributes are derived directly from the URDF parameters of the child joint associated with the edge, ensuring that local kinematic transformations are accurately preserved within the graph representation.

\subsubsection{Handling of Fixed Joints}

URDF models may contain fixed joints that do not introduce additional degrees of freedom. Instead of explicitly modeling fixed joints as graph nodes, we compose their rigid transformations into adjacent edges through transformation composition:
\begin{equation}
T_{ij} = \prod_k T_k^{\mathrm{fixed}},
\end{equation}
where $T_k^{\mathrm{fixed}}$ denotes the homogeneous transformation associated with an intermediate fixed joint.

Specifically, all intermediate fixed-joint transformations are merged into the edge attributes between neighboring actuated joints. This design avoids introducing redundant nodes while preserving the completeness of the kinematic structure, resulting in a compact and efficient graph representation.

\subsection{Graph Diffusion for Inverse Kinematics}

The proposed framework formulates inverse kinematics as a conditional graph diffusion process defined on the robot kinematic graph. Specifically, joint configurations are represented as graph node features, and the diffusion process is performed directly on the graph representation to model the conditional distribution
\begin{equation}
p(q \mid y),
\end{equation}
where $q$ denotes the robot joint configuration and $y$ represents the target end-effector pose.

At diffusion step $t$, the robot state is represented as a graph
\begin{equation}
G_t = (X_t, E),
\qquad
X_t = [q_t^{(1)}, q_t^{(2)}, \dots, q_t^{(N)}],
\end{equation}
where $X_t$ denotes the noisy node features at diffusion step $t$, and $E$ is the fixed graph topology defined by the robot kinematic structure. The diffusion process is applied only to the node features corresponding to joint states, while the graph connectivity remains unchanged throughout the denoising process.

\subsubsection{Forward Diffusion Process}

In the forward diffusion process, Gaussian noise is progressively added to the clean joint configuration. Given a clean configuration $q_0$, the noisy state at diffusion step $t$ is sampled as
\begin{equation}
q_t = \sqrt{\bar{\alpha}_t} q_0 + \sqrt{1-\bar{\alpha}_t}\epsilon,
\quad
\epsilon \sim \mathcal{N}(0, I),
\end{equation}
where $\bar{\alpha}_t$ denotes the cumulative noise schedule coefficient. This process gradually transforms the original joint configuration distribution into a Gaussian distribution while preserving the underlying graph structure.

\subsubsection{Conditional Reverse Denoising}

The reverse diffusion process learns to iteratively remove noise from the graph node features conditioned on the target task-space information. The denoising model is defined as
\begin{equation}
\epsilon_\theta(G_t, c, t),
\end{equation}
where $G_t$ is the noisy graph at diffusion step $t$, $t$ is the diffusion timestep, and $c$ denotes the conditioning information.

The condition consists of both target end-effector information and noisy forward-kinematics observations:
\begin{equation}
c = \{p, r, \hat{p}_t, \hat{r}_t\},
\end{equation}
where $p$ and $r$ denote the target end-effector position and orientation, respectively. The noisy end-effector observation is obtained by applying forward kinematics to the noisy joint configuration:
\begin{equation}
(\hat{p}_t, \hat{r}_t) = \mathrm{FK}(q_t),
\end{equation}
where $\hat{p}_t$ and $\hat{r}_t$ denote the noisy end-effector position and orientation at diffusion step $t$, respectively. These noisy task-space observations provide additional geometric guidance during the denoising process. For multi-branch robotic systems, such as dual-arm robots with torso, the conditioning is extended to include branch-specific target poses and noisy forward-kinematics observations for each end-effector.

Unlike conventional diffusion models operating on flattened joint representations, the proposed framework performs denoising directly on the robot kinematic graph through structure-aware message passing. For multi-branch robotic systems, the denoising process further incorporates staged graph reasoning, where torso and branch-specific subgraphs are modeled hierarchically to capture inter-branch coordination and structural dependencies. The predicted clean joint configuration can subsequently be reconstructed from the predicted noise following the standard diffusion formulation, enabling task-space supervision through forward kinematics.

\subsection{Structure-Aware Graph Convolution}
\label{sec:structure_aware_graph_conv}

Although graph neural networks are effective in modeling relationships among nodes, conventional graph convolution typically treats the graph as a homogeneous set of connections and does not explicitly exploit the hierarchical and branching structure inherent in robot kinematics. For articulated robots with torso structures and multiple kinematic branches, such as dual-arm systems, modeling these structural dependencies is crucial for coordinated motion generation and inverse kinematics reasoning.

\begin{figure}[!t]
    \centering
    \includegraphics[width=\linewidth]{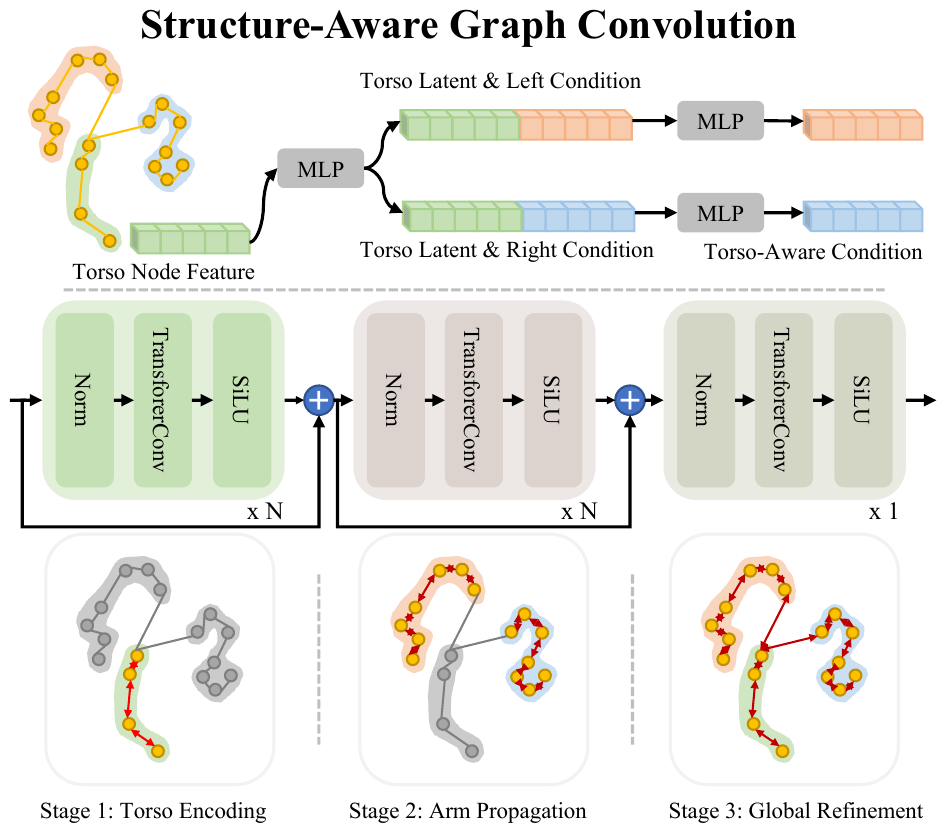}
    \caption{\textbf{Structure-Aware Graph Convolution.}
    We perform stage-wise message passing over the robot kinematic graph derived from the URDF.
    Given node features encoding joint states and structural attributes, the model applies a sequence of structure-aware graph convolution blocks with residual connections.
    In Stage~1, torso nodes are updated to capture global context and produce a shared latent representation $z_{\text{torso}}$.
    In Stage~2, torso information is propagated to the left and right arms through conditioned message passing, enabling coordinated feature updates across kinematic branches.
    Finally, Stage~3 performs global refinement over the entire graph to improve consistency across all joints.
    }
    \label{fig:structure_aware_graph_conv}
\end{figure}

To address this limitation, we propose a structure-aware graph convolution framework, as illustrated in Fig.~\ref{fig:structure_aware_graph_conv}. The proposed approach combines attention-based graph message passing, conditional feature modulation, and stage-wise hierarchical reasoning to explicitly model multi-branch kinematic dependencies.

\subsubsection{Transformer-based Graph Convolution}

At each message passing layer, we adopt an attention-based graph convolution operator based on TransformerConv. Let $h_i^{(l)}$ denote the feature of node $i$ at layer $l$. The node update is defined as
\begin{equation}
h_i^{(l+1)}
=
W_1 h_i^{(l)}
+
\sum_{j \in \mathcal{N}(i)}
\alpha_{ij}^{(l)}
\left(
W_2 h_j^{(l)}
+
W_e e_{ij}
\right),
\end{equation}
where $\mathcal{N}(i)$ denotes the neighborhood of node $i$, $W_1$, $W_2$, and $W_e$ are learnable parameters, and $e_{ij}$ represents the edge feature between nodes $i$ and $j$. The attention coefficients are computed as
\begin{equation}
\alpha_{ij}^{(l)}
=
\mathrm{softmax}_j
\left(
\frac{
(W_q h_i^{(l)})^\top
(W_k h_j^{(l)} + W_e e_{ij})
}{
\sqrt{d}
}
\right),
\end{equation}
where $W_q$ and $W_k$ are learnable projection matrices, and $d$ denotes the feature dimension. This formulation enables the model to adaptively aggregate neighboring information based on both node features and local kinematic relationships encoded in the edge attributes.

\subsubsection{Conditional Feature Modulation}

To incorporate task-space conditioning information into the graph denoising process, we adopt a Feature-wise Linear Modulation (FiLM)~\cite{perez2018film} mechanism for conditional feature transformation. Given a conditioning vector $c$, a multi-layer perceptron predicts scaling and bias parameters:
\begin{equation}
[\gamma, \beta] = \mathrm{MLP}(c),
\end{equation}
and the node features are normalized and then modulated as
\begin{equation}
\tilde{h}_i^{(l)}
=
\gamma
\odot
\mathrm{Norm}(h_i^{(l)})
+
\beta,
\end{equation}
where $\odot$ denotes element-wise multiplication. This modulation mechanism enables the graph convolution process to adaptively adjust feature propagation according to the diffusion timestep, target end-effector pose, and noisy forward-kinematics observations.

\subsubsection{Stage-wise Graph Convolution}

Built upon the above graph convolution operator, we further design a structure-aware stage-wise reasoning framework to explicitly model hierarchical dependencies in multi-branch robotic systems. Instead of performing message passing uniformly over the entire graph, the proposed framework decomposes the robot graph into structure-aware subgraphs:
\begin{equation}
G_{\mathrm{torso}} \subseteq G,
\qquad
G_{\mathrm{arms}} \subseteq G,
\end{equation}
where $G_{\mathrm{torso}}$ contains torso-related joints and $G_{\mathrm{arms}}$ contains arm-related branches.

\begin{itemize}
    \item \textbf{Stage 1: Torso Modeling.}
    Message passing is performed only on the torso subgraph:
    \begin{equation}
    H_{\mathrm{torso}}
    =
    \mathrm{GNN}(G_{\mathrm{torso}}),
    \end{equation}
    where $H_{\mathrm{torso}}$ denotes the torso node features after graph convolution. The torso features are subsequently aggregated into a shared latent representation:
    \begin{equation}
    z_{\mathrm{torso}}
    =
    f_{\mathrm{proj}}(H_{\mathrm{torso}}),
    \end{equation}
    where $f_{\mathrm{proj}}(\cdot)$ denotes a learnable projection function. The resulting latent representation encodes global robot pose information and serves as a coordination condition for subsequent branch-wise reasoning.

    \item \textbf{Stage 2: Branch-aware Arm Reasoning.}
    Message passing is performed on arm-related subgraphs while conditioning on the shared torso latent representation $z_{\mathrm{torso}}$. Specifically, left-arm and right-arm branches are modulated independently through branch-specific conditioning functions:
    \begin{equation}
    \tilde{h}_{i,\mathrm{left}}
    =
    \mathrm{FiLM}(h_i, c_{\mathrm{left}}, z_{\mathrm{torso}}),
    \end{equation}
    \begin{equation}
    \tilde{h}_{i,\mathrm{right}}
    =
    \mathrm{FiLM}(h_i, c_{\mathrm{right}}, z_{\mathrm{torso}}),
    \end{equation}
    where $c_{\mathrm{left}}$ and $c_{\mathrm{right}}$ denote the branch-specific conditioning information. This design enables coordinated information propagation across different kinematic branches while preserving branch-specific motion characteristics.

    \item \textbf{Stage 3: Global Refinement.}
    Finally, a refinement stage performs message passing over the complete graph:
    \begin{equation}
    H_{\mathrm{refine}}
    =
    \mathrm{GNN}(G),
    \end{equation}
    allowing information from different branches to be jointly integrated. This refinement stage improves inter-branch consistency and produces globally coherent joint configurations for inverse kinematics prediction.

\end{itemize}

\subsection{Conditioning Mechanism}
\label{sec:conditioning_mechanism}

To guide the diffusion process toward target-driven and physically consistent solutions, we design a hierarchical conditioning mechanism that incorporates both task-level objectives and state-dependent geometric feedback. Specifically, the conditioning consists of three components: diffusion timestep encoding, target end-effector pose encoding, and noisy forward kinematics feedback. These conditioning signals are encoded into latent representations and injected into the graph diffusion network through the FiLM-based feature modulation mechanism described in Sec.~\ref{sec:structure_aware_graph_conv}.

\subsubsection{Timestep Encoding}

The diffusion timestep $t$ indicates the current stage of the denoising process. We encode the timestep using a positional embedding followed by a multi-layer perceptron:
\begin{equation}
c_t = \phi_t(t),
\end{equation}
where $\phi_t(\cdot)$ denotes the timestep encoding function. This representation provides global information regarding the current diffusion stage.

\subsubsection{Target Pose Encoding}

Given a target end-effector pose
\begin{equation}
y = (p, r),
\end{equation}
where $p \in \mathbb{R}^3$ denotes the target position and $r \in \mathbb{R}^6$ denotes the target orientation represented in rot6d~\cite{zhou2019continuity} form, the target pose condition is encoded as
\begin{equation}
c_{\mathrm{pose}}
=
[
\phi_p(p)
\|
\phi_r(r)
].
\end{equation}
where $\phi_p(\cdot)$ and $\phi_r(\cdot)$ denote learnable encoders for position and orientation, respectively. For multi-branch robotic systems, branch-specific target pose conditions are constructed independently for each end-effector.

\subsubsection{Noisy Forward Kinematics Feedback}

To provide state-dependent geometric guidance during denoising, we compute the forward kinematics of the noisy joint configuration:
\begin{equation}
\hat{y}_t = \mathrm{FK}(q_t).
\end{equation}

The noisy end-effector pose is decomposed into position and orientation:
\begin{equation}
\hat{y}_t = (\hat{p}_t, \hat{u}_t),
\end{equation}
where $\hat{p}_t \in \mathbb{R}^3$ denotes the noisy end-effector position and $\hat{u}_t \in \mathbb{R}^4$ denotes the quaternion representation of orientation. The noisy FK condition is then encoded as
\begin{equation}
c_{\mathrm{fk}}
=
[
\phi_{\hat{p}}(\hat{p}_t)
\|
\phi_{\hat{u}}(\hat{u}_t)
].
\end{equation}

This conditioning provides explicit geometric feedback regarding the current task-space state induced by the noisy joint configuration, enabling the denoising process to reason jointly in both joint space and task space. For multi-branch robotic systems, branch-specific noisy FK observations are computed independently for each end-effector branch.

\subsubsection{Hierarchical Conditioning Formulation}

For single-arm robots, the overall conditioning vector is constructed as
\begin{equation}
c
=
[
c_t
\|
c_{\mathrm{pose}}
\|
c_{\mathrm{fk}}
].
\end{equation}
For multi-branch robotic systems, branch-specific conditioning vectors are constructed independently:
\begin{equation}
c^{L}
=
[
c_t
\|
c_{\mathrm{pose}}^{L}
\|
c_{\mathrm{fk}}^{L}
],
\end{equation}
\begin{equation}
c^{R}
=
[
c_t
\|
c_{\mathrm{pose}}^{R}
\|
c_{\mathrm{fk}}^{R}
].
\end{equation}

To further coordinate different kinematic branches, we incorporate a global latent representation extracted from the torso subgraph:
\begin{equation}
z_{\mathrm{torso}}
=
f_{\mathrm{proj}}(H_{\mathrm{torso}}),
\end{equation}
where $H_{\mathrm{torso}}$ denotes the torso feature representation obtained from Stage~1 of the structure-aware graph reasoning process. The final branch-aware conditioning vectors are obtained by fusing local branch conditions with the global torso latent:
\begin{equation}
\tilde{c}^{L}
=
[
c^{L}
\|
z_{\mathrm{torso}}
],
\qquad
\tilde{c}^{R}
=
[
c^{R}
\|
z_{\mathrm{torso}}
].
\end{equation} This hierarchical conditioning strategy enables both local branch-specific control and global coordination across articulated kinematic branches.

\subsubsection{Stage-wise Conditioning Usage}

The proposed conditioning mechanism is injected hierarchically throughout the structure-aware graph reasoning process. Global conditioning information is first utilized during torso-level reasoning to extract the shared latent representation $z_{\mathrm{torso}}$. Subsequently, branch-specific conditioning vectors are incorporated into left-arm and right-arm message passing for branch-aware denoising. Finally, global conditioning and torso-aware structural information are jointly integrated during the full-graph refinement stage to improve inter-branch consistency and globally coherent joint generation.

\subsection{Training Objective}

The training objective consists of two components: a standard diffusion noise prediction loss for learning the reverse denoising process, and a forward kinematics loss that enforces task-space consistency. The former enables stable denoising in the joint space, while the latter explicitly supervises the generated joint configurations in the end-effector space.

\subsubsection{Noise Prediction Loss}

Following the standard diffusion training formulation, the model is trained to predict the injected Gaussian noise. Given a noisy joint configuration $q_t$ at diffusion step $t$, the denoising model predicts the noise as
\begin{equation}
\epsilon_\theta(G_t, c, t),
\end{equation}
where $G_t$ denotes the noisy graph representation and $c$ represents the conditioning information, the noise prediction objective is defined as
\begin{equation}
L_{\mathrm{noise}}
=
\mathbb{E}_{t,q_0,\epsilon}
\left[
\left\|
\epsilon -
\epsilon_\theta(G_t, c, t)
\right\|^2
\right],
\end{equation}
this objective enables the model to learn the reverse diffusion process and progressively recover clean joint configurations from noisy graph states.

\subsubsection{Reconstruction of Clean Joint Configuration}

Since the diffusion model predicts noise rather than joint values directly, the clean joint configuration is reconstructed from the noisy input using the standard diffusion formulation:
\begin{equation}
\hat{q}_0
=
\frac{
q_t -
\sqrt{1-\bar{\alpha}_t}
\,
\epsilon_\theta(G_t,c,t)
}{
\sqrt{\bar{\alpha}_t}
},
\end{equation}
where $\bar{\alpha}_t$ denotes the cumulative product of the diffusion noise schedule up to timestep $t$. Before forward kinematics evaluation, the reconstructed joint configurations are transformed back to the original joint space through denormalization.

\subsubsection{Forward Kinematics Supervision}

Although the diffusion objective models the distribution of joint configurations, it does not explicitly enforce consistency in task space. In inverse kinematics, multiple joint configurations may satisfy similar task-space constraints, while small errors in joint space can still lead to large deviations in end-effector space. To address this issue, we introduce a forward kinematics supervision loss. Given the reconstructed joint configuration $\hat{q}_0$, the corresponding end-effector pose is computed as
\begin{equation}
\hat{y}_0 = \mathrm{FK}(\hat{q}_0).
\end{equation}
Let the target pose be
\begin{equation}
y = (p, r),
\end{equation}
where $p$ denotes the target position and $r$ denotes the target orientation represented in rot6d form. For rotation supervision, the target orientation is converted into quaternion representation $u$. The FK supervision loss is defined as
\begin{equation}
L_{\mathrm{FK}}
=
L_{\mathrm{pos}}
+
\lambda_{\mathrm{rot}}
L_{\mathrm{rot}},
\end{equation}
where $L_{\mathrm{pos}}$ and $L_{\mathrm{rot}}$ denote the position and orientation losses, respectively. The position loss is computed as
\begin{equation}
L_{\mathrm{pos}}
=
\left\|
\hat{p}_0 - p
\right\|^2,
\end{equation}
where $\hat{p}_0$ denotes the predicted end-effector position.

For orientation supervision, let $\hat{u}_0$ and $u$ denote the predicted and target unit quaternions, respectively. The rotation loss is defined as
\begin{equation}
L_{\mathrm{rot}}
=
L_{\mathrm{quat}}(\hat{u}_0, u),
\end{equation}
where $L_{\mathrm{quat}}(\cdot)$ denotes a quaternion geodesic loss measuring orientation discrepancy while remaining invariant to quaternion sign ambiguity. For multi-branch robotic systems, the FK supervision loss is computed independently for each end-effector branch and summed together during optimization.

\subsubsection{Late-step FK Supervision}

Instead of applying FK supervision throughout the entire diffusion trajectory, the FK loss is activated only during the later stages of the denoising process. At early diffusion stages, the joint configurations are heavily corrupted by Gaussian noise, and the resulting forward kinematics observations are not geometrically meaningful. Enforcing task-space constraints at this stage may interfere with learning the fundamental denoising dynamics.

As the denoising process progresses, the reconstructed joint configurations gradually approach the solution manifold, making the corresponding forward kinematics observations increasingly reliable. Applying FK supervision during these later denoising stages provides effective geometric guidance toward task-space consistency.
Let $\mathbb{I}_{\mathrm{FK}}(t)$ denote an indicator function defined as
\begin{equation}
\mathbb{I}_{\mathrm{FK}}(t)
=
\mathbb{I}(t < \tau_{\mathrm{FK}}),
\end{equation}
where $\tau_{\mathrm{FK}}$ denotes the FK supervision threshold. The final training objective is therefore defined as
\begin{equation}
L
=
L_{\mathrm{noise}}
+
\lambda_{\mathrm{FK}}
\,
\mathbb{I}_{\mathrm{FK}}(t)
\,
L_{\mathrm{FK}},
\end{equation}
where $\lambda_{\mathrm{FK}}$ controls the contribution of task-space supervision during training. The overall training procedure of the proposed GraphDiff-IK framework is summarized in Algorithm~\ref{alg:graphdiffik_learning}.

\begin{algorithm}[!t]
\caption{GraphDiff-IK--Learning}
\label{alg:graphdiffik_learning}
\begin{algorithmic}[1]
\Require IK dataset $\mathcal{D}=\{(q_0,y)\}$, robot graph $G=(V,E)$, denoising network $\epsilon_\theta$, noise schedule $\{\bar{\alpha}_t\}_{t=1}^{T}$, learning rate $\eta$, FK threshold $\tau_{\mathrm{FK}}$, loss weights $\lambda_{\mathrm{FK}},\lambda_{\mathrm{rot}}$
\Ensure Optimized parameters $\theta$
\While{not converged}
    \State $(q_0,y)\sim\mathcal{D}$, \quad $t\sim\mathcal{U}(1,T)$, \quad $\epsilon\sim\mathcal{N}(0,I)$
    \State $q_t=\sqrt{\bar{\alpha}_t}q_0+\sqrt{1-\bar{\alpha}_t}\epsilon$ \Comment{forward diffusion}
    \State $G_t=(X_t,E)$ \Comment{construct noisy kinematic graph}
    \State $\hat{y}_t=\mathrm{FK}(q_t)$ \Comment{state-dependent task-space feedback}
    \State $c=\mathcal{C}(t,y,\hat{y}_t)$ \Comment{hierarchical conditioning}
    \State $\hat{\epsilon}=\epsilon_\theta(G_t,c,t)$ \Comment{structure-aware graph denoising}
    \State $\hat{q}_0=\dfrac{q_t-\sqrt{1-\bar{\alpha}_t}\hat{\epsilon}}{\sqrt{\bar{\alpha}_t}}$
    \State $\hat{y}_0=\mathrm{FK}(\hat{q}_0)$
    \State $L_{\mathrm{noise}}=\|\epsilon-\hat{\epsilon}\|_2^2$
    \State $L_{\mathrm{FK}}=\|\hat{p}_0-p\|_2^2+\lambda_{\mathrm{rot}}L_{\mathrm{quat}}(\hat{u}_0,u)$
    \State $L=L_{\mathrm{noise}}+\lambda_{\mathrm{FK}}\mathbb{I}(t<\tau_{\mathrm{FK}})L_{\mathrm{FK}}$
    \State $\theta\leftarrow\theta-\eta\nabla_\theta L$
\EndWhile
\end{algorithmic}
\end{algorithm}

\subsection{Inference and Deployment}

\subsubsection{Inference Process}

At inference time, the model generates joint configurations by reversing the diffusion process. Starting from an initial Gaussian noise sample:
\begin{equation}
q_T \sim \mathcal{N}(0, I),
\end{equation}
the model iteratively performs reverse denoising from timestep $t=T$ to $t=0$. At each denoising step, the model predicts the injected noise conditioned on the robot graph state and the conditioning information:
\begin{equation}
\epsilon_\theta(G_t, c, t),
\end{equation}
where $G_t$ denotes the noisy robot graph and $c$ represents the hierarchical conditioning information described in Sec.~\ref{sec:conditioning_mechanism}.

The noisy joint configuration is then updated through the denoising diffusion implicit models (DDIM)~\cite{song2021ddim} reverse diffusion process:
\begin{equation}
q_{t-1}
=
\mathrm{DDIM}
(
q_t,
\epsilon_\theta(G_t,c,t),
t
),
\end{equation}
through iterative denoising, the joint configuration progressively converges toward the target-conditioned solution manifold. After completing the reverse diffusion trajectory, the final output $\hat{q}_0$ is obtained as the predicted joint configuration satisfying the desired end-effector constraints. In practice, the reverse diffusion process can be performed using a reduced number of DDIM sampling steps, enabling efficient inference while maintaining accurate task-space consistency. This allows a practical trade-off between computational cost and solution quality, while supporting efficient inference for both single-arm and multi-branch robotic systems. The complete inference procedure of the proposed framework is summarized in Algorithm~\ref{alg:graphdiffik_inference}.

\subsubsection{Deployment and Multi-modal Generation}

The proposed framework can be directly deployed as a learned inverse kinematics solver for articulated robotic systems, generating feasible joint configurations without requiring iterative optimization or analytical inverse kinematics derivations. Unlike deterministic regression-based methods, the proposed generative framework can sample multiple valid joint configurations for the same target pose, which is particularly important for redundant and multi-branch robotic systems. Furthermore, since the framework operates directly on robot kinematic graphs, the same formulation naturally generalizes across diverse robot morphologies, including single-arm manipulators, dual-arm systems, and articulated robots with torso or waist structures.

\begin{algorithm}[!t]
\caption{GraphDiff-IK--Inference}
\label{alg:graphdiffik_inference}
\begin{algorithmic}[1]
\Require Robot graph $G=(V,E)$, target pose $y$, trained denoising network $\epsilon_\theta$, DDIM sampling schedule, sampling steps $T$
\Ensure Predicted joint configuration $\hat{q}_0$
\State $q_T\sim\mathcal{N}(0,I)$ \Comment{initialize graph node noise}
\For{$t=T,\ldots,1$}
    \State $G_t=(X_t,E)$ \Comment{construct noisy kinematic graph}
    \State $\hat{y}_t=\mathrm{FK}(q_t)$
    \State $c=\mathcal{C}(t,y,\hat{y}_t)$
    \State $\hat{\epsilon}_t=\epsilon_\theta(G_t,c,t)$ \Comment{structure-aware graph denoising}
    \State $q_{t-1}=\mathrm{DDIM}(q_t,\hat{\epsilon}_t,t)$
\EndFor
\State \Return $\hat{q}_0=q_0$
\end{algorithmic}
\end{algorithm}


\section{EXPERIMENTS}

\subsection{Experimental Setup}

\subsubsection{Robotic Platforms}

\begin{figure}[!t]
    \centering
    \includegraphics[width=\linewidth]{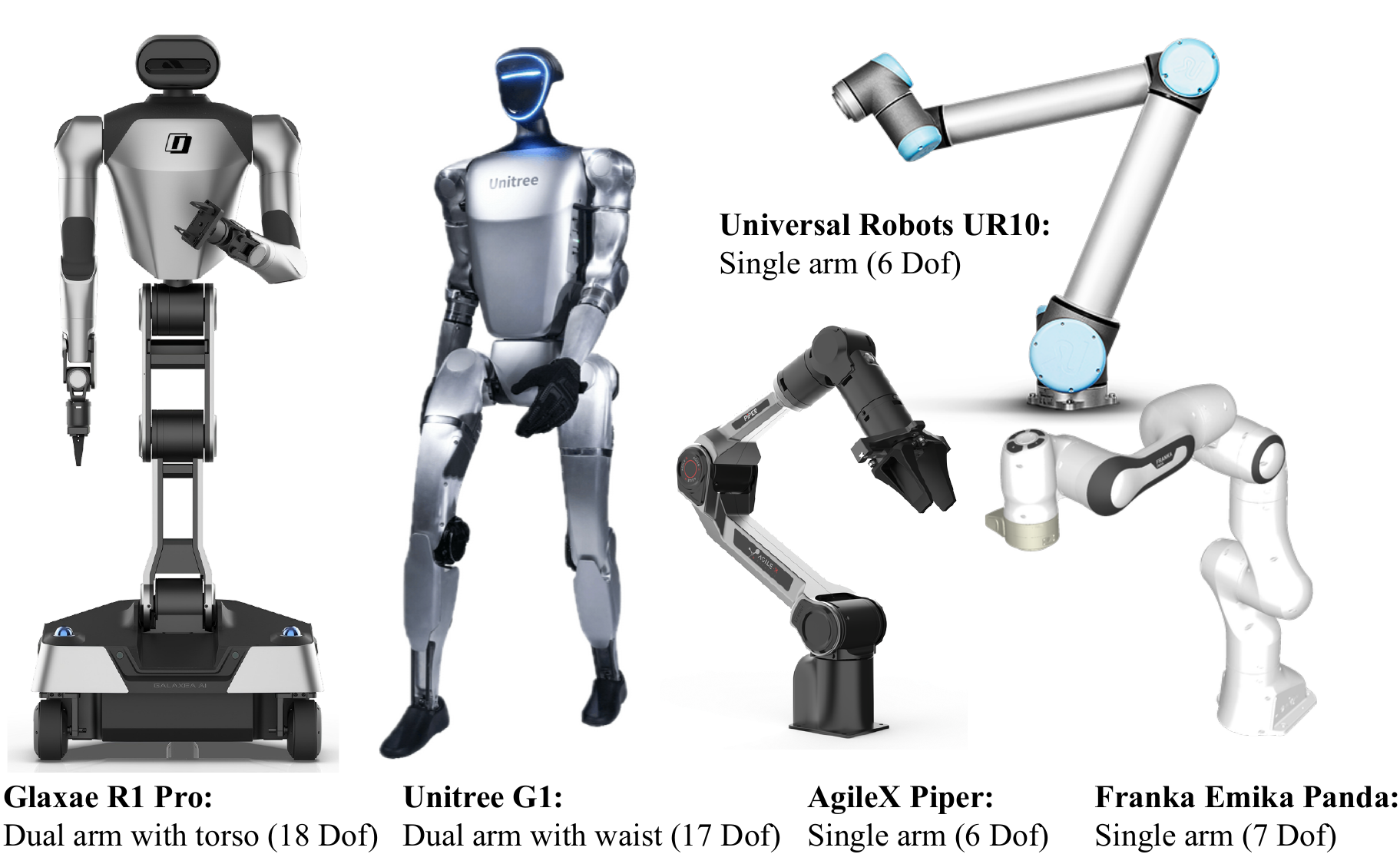}
    \caption{\textbf{Robot platforms used in our experiments.} The evaluated platforms include fixed-base single-arm manipulators, dual-arm systems with torso articulation, and humanoid robots with waist coupling, covering diverse kinematic structures and degrees of freedom.
}
    \label{fig:robots}
\end{figure}

To evaluate the proposed GraphDiff-IK framework across diverse kinematic structures, we consider multiple robotic platforms with varying degrees of freedom and topological complexity, including both single-arm manipulators and multi-branch robotic systems, as illustrated in Fig.~\ref{fig:robots}. The evaluated single-arm manipulators include the Franka Emika Panda (7-DoF), Universal Robots UR10 (6-DoF), and AgileX Piper (6-DoF). These robots cover both redundant and non-redundant kinematic configurations. In particular, 6-DoF manipulators are typically non-redundant for end-effector pose control, while 7-DoF manipulators exhibit kinematic redundancy and admit multiple valid inverse kinematics solutions for the same target pose. To further evaluate scalability on more complex articulated systems, we additionally consider the Unitree G1 humanoid robot (14-DoF upper body configuration) and the Galaxea R1 Pro dual-arm platform (18-DoF with torso articulation). Compared to fixed-base single-arm manipulators, these systems exhibit substantially more complex kinematic structures due to coordinated multi-branch articulation and increased degrees of freedom. For all platforms, robot kinematics are represented using the unified kinematic graph formulation described in Section~III-B. The same network architecture and training pipeline are applied across all robots without robot-specific modifications, demonstrating the generalization capability of GraphDiff-IK across diverse robot morphologies and kinematic structures.

\subsubsection{Dataset Generation}

For each robot platform, inverse kinematics datasets are generated using FK-based sampling. Specifically, joint configurations are randomly sampled within the corresponding joint limits, and forward kinematics is applied to compute the associated end-effector poses. Each training sample therefore consists of a joint configuration and its corresponding target end-effector pose. This formulation enables efficient large-scale data generation without requiring manually annotated IK solutions. For single-arm manipulators, the target pose consists of the end-effector position and orientation of the corresponding manipulator. For dual-arm systems, target poses are generated independently for both arms while preserving the shared torso or waist articulation.

\subsubsection{Evaluation Metrics}

We evaluate inverse kinematics performance using both end-effector position error and orientation error. Position error is computed as the Euclidean distance between the predicted and target end-effector positions and is reported in millimeters. Orientation error is measured as the angular distance between predicted and target end-effector orientations and is reported in degrees. Following the evaluation protocol used in IKNet~\cite{iknet}, all quantitative results are reported over 100 randomly sampled target poses. For dual-arm robotic systems, evaluation metrics are reported separately for the left and right arms. In addition to quantitative evaluation, we further analyze the proposed method through qualitative visualization of workspace coverage, multi-solution inverse kinematics generation, and diffusion denoising trajectories.

\subsubsection{Implementation Details}

The proposed framework is implemented using PyTorch~\cite{pytorch} and PyTorch Geometric~\cite{pytorch_geometric}. The graph diffusion model is trained using the DDIM scheduler with 100 diffusion timesteps and epsilon prediction objective. During training, Gaussian noise is progressively added to joint configurations, and the network is trained to predict the injected noise conditioned on target end-effector poses. Unless otherwise specified, the model uses a hidden feature dimension of 512 and timestep embedding dimension of 256. The network is optimized using AdamW with a learning rate of $1\times10^{-4}$ and weight decay of $1\times10^{-4}$. Gradient clipping and cosine learning rate scheduling with warmup are applied during training. In addition to the diffusion noise prediction objective, FK-based geometric supervision is introduced during later denoising stages to improve end-effector consistency.

\subsection{Comparison with Baseline Methods}

We compare the proposed GraphDiff-IK framework with several representative inverse kinematics approaches, including regression-based methods (MLP~\cite{rumelhart1986learning}, Transformer~\cite{transformer}, and GNN~\cite{kipf2016semi}), as well as recent learning-based and generative IK methods, including IKNet~\cite{iknet}, IKFlow~\cite{ikflow}, and GGIK~\cite{ggik}. Experiments are conducted on the Franka Emika Panda and Universal Robots UR10 platforms and quantitative results are summarized in Table~\ref{tab:comparison}.

\begin{table*}[t]
\caption{Quantitative comparison of inverse kinematics performance between GraphDiff-IK and baseline methods on the Franka Emika Panda and Universal Robots UR10. Position error is reported in millimeters, and rotation error is reported in degrees. Lower values indicate better performance. IKNet only reports position-related metrics and does not provide orientation prediction results.}
\label{tab:comparison}
\centering
\setlength{\tabcolsep}{2pt}
\renewcommand{\arraystretch}{1.0}
\begin{tabular}{l ccc ccc ccc ccc}
\toprule
\multirow{4}{*}{\textbf{Method}} 
& \multicolumn{6}{c}{\textbf{Franka Emika Panda}} 
& \multicolumn{6}{c}{\textbf{Universal Robots UR10}} \\
\cmidrule(lr){2-7} \cmidrule(lr){8-13}
& \multicolumn{3}{c}{\textbf{Error Pos (mm)}} 
& \multicolumn{3}{c}{\textbf{Error Rot (deg)}} 
& \multicolumn{3}{c}{\textbf{Error Pos (mm)}} 
& \multicolumn{3}{c}{\textbf{Error Rot (deg)}} \\
\cmidrule(lr){2-4} \cmidrule(lr){5-7} \cmidrule(lr){8-10} \cmidrule(lr){11-13}
& Mean$\pm$Std & Min & Max & Mean$\pm$Std & Min & Max
& Mean$\pm$Std & Min & Max & Mean$\pm$Std & Min & Max \\
\midrule
Transformer~\cite{transformer} & 29.84 $\pm$ 25.89 & 2.92 & 146.25 & 6.47 $\pm$4.83 & 0.73 & 22.55 & 29.30$\pm$55.50 & 1.62 & 410.62 & 2.80$\pm$3.50 & 0.18 & 21.03 \\
GNN~\cite{kipf2016semi}     & 339.25$\pm$205.02 & 44.28 & 900.38 & 72.47$\pm$39.78 & 12.07 & 177.31 & 464.09$\pm$278.39 & 46.03 & 1459.64 & 68.45 $\pm$ 38.93 & 11.78 & 177.53 \\
MLP~\cite{rumelhart1986learning}    & 496.54 $\pm$ 286.13 & 23.50 & 1458.82 & 90.76 $\pm$ 43.23 & 12.14 & 179.90 & 673.18 $\pm$ 404.92 & 73.07 & 1732.49 & 111.38 $\pm$ 46.58 & 14.59 & 179.88 \\
GGIK~\cite{ggik}   & 6.17$\pm$2.71 & 2.28 & 12.80 & \textbf{0.59$\pm$0.25} & 0.21 & 1.15 & 6.85$\pm$1.77 & 4.14 & 10.93 & \textbf{0.34$\pm$0.09} & 0.18 & 0.55 \\
IKNet~\cite{iknet}  & 36.17$\pm$6.79 & 0.18 & 570.60 & - & - & - & 125.00$\pm$19.68 & 2.31 & 1786.74 & - & - & - \\
\midrule
\textbf{GraphDiff-IK (Ours)}   & \textbf{3.47$\pm$2.47} & 0.18 & 12.40 & 0.87$\pm$0.69 & 0.10 & 4.26
       & \textbf{3.99$\pm$2.43} & 0.41 & 17.44 & 1.05$\pm$0.79 & 0.08 & 5.09 \\
\bottomrule
\end{tabular}
\end{table*}

Overall, the proposed GraphDiff-IK framework achieves the best positional accuracy across both robot platforms while maintaining competitive orientation accuracy. In particular, GraphDiff-IK achieves a mean position error of 3.47 mm on the Franka Emika Panda and 3.99 mm on the Universal Robots UR10, outperforming all compared baseline methods in positional accuracy. Compared with GGIK, which achieves the strongest baseline performance among existing methods, GraphDiff-IK reduces the mean position error from 6.17 mm to 3.47 mm on Franka and from 6.85 mm to 3.99 mm on UR10.

GraphDiff-IK significantly outperforms regression-based methods, including MLP, Transformer, and GNN models, demonstrating the advantage of diffusion-based structured generation for inverse kinematics problems. The large errors observed in direct regression approaches suggest that learning a deterministic mapping from end-effector poses to joint configurations is insufficient for capturing the complex and multi-modal nature of inverse kinematics.

Although GGIK achieves slightly lower orientation error, GraphDiff-IK maintains substantially better positional accuracy and more stable overall performance across different robot morphologies. In addition, GraphDiff-IK exhibits relatively small variance and bounded maximum errors compared to several baseline methods, indicating improved robustness and stability during inference. These results demonstrate that modeling inverse kinematics as a conditional graph diffusion process provides a strong and flexible formulation for generating accurate joint configurations while preserving robot kinematic structure.

\subsection{Generalization Across Robot Morphologies}

To evaluate the generalization capability of the proposed framework across different robot morphologies, we conduct experiments on robotic systems with varying degrees of freedom and kinematic topologies, including single-arm manipulators, dual-arm systems, and humanoid robots. Quantitative results are summarized in Table~\ref{tab:generalization}.

\begin{table*}[t]
\caption{Generalization performance of GraphDiff-IK across diverse robot morphologies, including single-arm manipulators and multi-branch robotic systems. Position error is reported in millimeters, and rotation error is reported in degrees, where lower values indicate better performance. For dual-arm robots, results are reported separately for the left and right arms.}
\label{tab:generalization}
\centering
\setlength{\tabcolsep}{2pt}
\renewcommand{\arraystretch}{1.0}
\begin{tabular}{l c l ccc ccc ccc ccc}
\toprule
\multirow[c]{4}{*}{\textbf{Robot}} 
& \multirow[c]{4}{*}{\textbf{DoF}} 
& \multirow[c]{4}{*}{\textbf{Description}} 
& \multicolumn{6}{c}{\textbf{Single Arm / Left Arm}} 
& \multicolumn{6}{c}{\textbf{Right Arm}} \\
\cmidrule(lr){4-9} \cmidrule(lr){10-15}
& & 
& \multicolumn{3}{c}{\textbf{Error Pos (mm)}} 
& \multicolumn{3}{c}{\textbf{Error Rot (deg)}}
& \multicolumn{3}{c}{\textbf{Error Pos (mm)}} 
& \multicolumn{3}{c}{\textbf{Error Rot (deg)}} \\
\cmidrule(lr){4-6} \cmidrule(lr){7-9} \cmidrule(lr){10-12} \cmidrule(lr){13-15}
& & 
& Mean$\pm$Std & Min & Max & Mean$\pm$Std & Min & Max
& Mean$\pm$Std & Min & Max & Mean$\pm$Std & Min & Max \\
\midrule
AgileX Piper & 6 & Single arm & 0.65$\pm$0.64 & 0.03 & 4.37 & 0.29$\pm$0.22 & 0.02 & 1.36 & - & - & - & - & - & - \\

Franka Emika Panda    & 7  & Single arm           & 3.47$\pm$2.47 & 0.18 & 12.40 & 0.87$\pm$0.69 & 0.10 & 4.26 & - & - & - & - & - & -  \\

Universal Robots UR10 & 6  & Single arm           & 3.99$\pm$2.43 & 0.41 & 17.44 & 1.05$\pm$0.79 & 0.08 & 5.09 & - & - & - & - & - & - \\

Unitree G1 & 14 & Dual arm with waist  & 0.82$\pm$0.68 & 0.05 & 3.88 & 0.33$\pm$0.24 & 0.03 & 1.20 & 0.73$\pm$0.46 & 0.08 & 2.52 & 0.32$\pm$0.18 & 0.04 & 0.85 \\

Galaxea R1 Pro & 18 & Dual arm with torso  & 5.99$\pm$4.48 & 0.54 & 23.81 & 0.62$\pm$0.40 & 0.03 & 2.39 & 4.59$\pm$2.95 & 0.85 & 14.73 & 0.55$\pm$0.29 & 0.11 & 1.26 \\
\bottomrule
\end{tabular}
\end{table*}

Overall, GraphDiff-IK achieves stable and accurate inverse kinematics performance across all evaluated robot platforms, despite significant differences in degrees of freedom and kinematic topology. The evaluated systems range from 6-DoF fixed-base manipulators to 18-DoF dual-arm articulated systems, demonstrating the scalability of the proposed method across increasingly complex robot morphologies.

For single-arm manipulators, including AgileX Piper, Franka Emika Panda, and Universal Robots UR10, the proposed method achieves low positional and rotational errors across both redundant and non-redundant configurations. In particular, GraphDiff-IK achieves mean position errors of 0.65 mm on AgileX Piper, 3.47 mm on Franka, and 3.99 mm on UR10, while maintaining orientation errors below 1.1 degrees across all evaluated platforms. These results demonstrate stable inverse kinematics generation across different manipulator structures and redundancy properties.

For more complex articulated systems, including the Unitree G1 humanoid robot and the Galaxea R1 Pro dual-arm with torso platform, GraphDiff-IK continues to achieve accurate inverse kinematics solutions for coordinated multi-branch motion. On Unitree G1, the proposed method achieves mean position errors below 1 mm for both arms, while on the 18-DoF Galaxea R1 Pro platform, the method maintains mean position errors of 5.99 mm and 4.59 mm for the left and right arms, respectively. Despite the increased kinematic complexity introduced by dual-arm articulation and higher degrees of freedom, the proposed graph-based diffusion formulation maintains stable performance and low inference error.

\begin{figure*}[!t]
    \centering
    \includegraphics[width=0.99\linewidth]{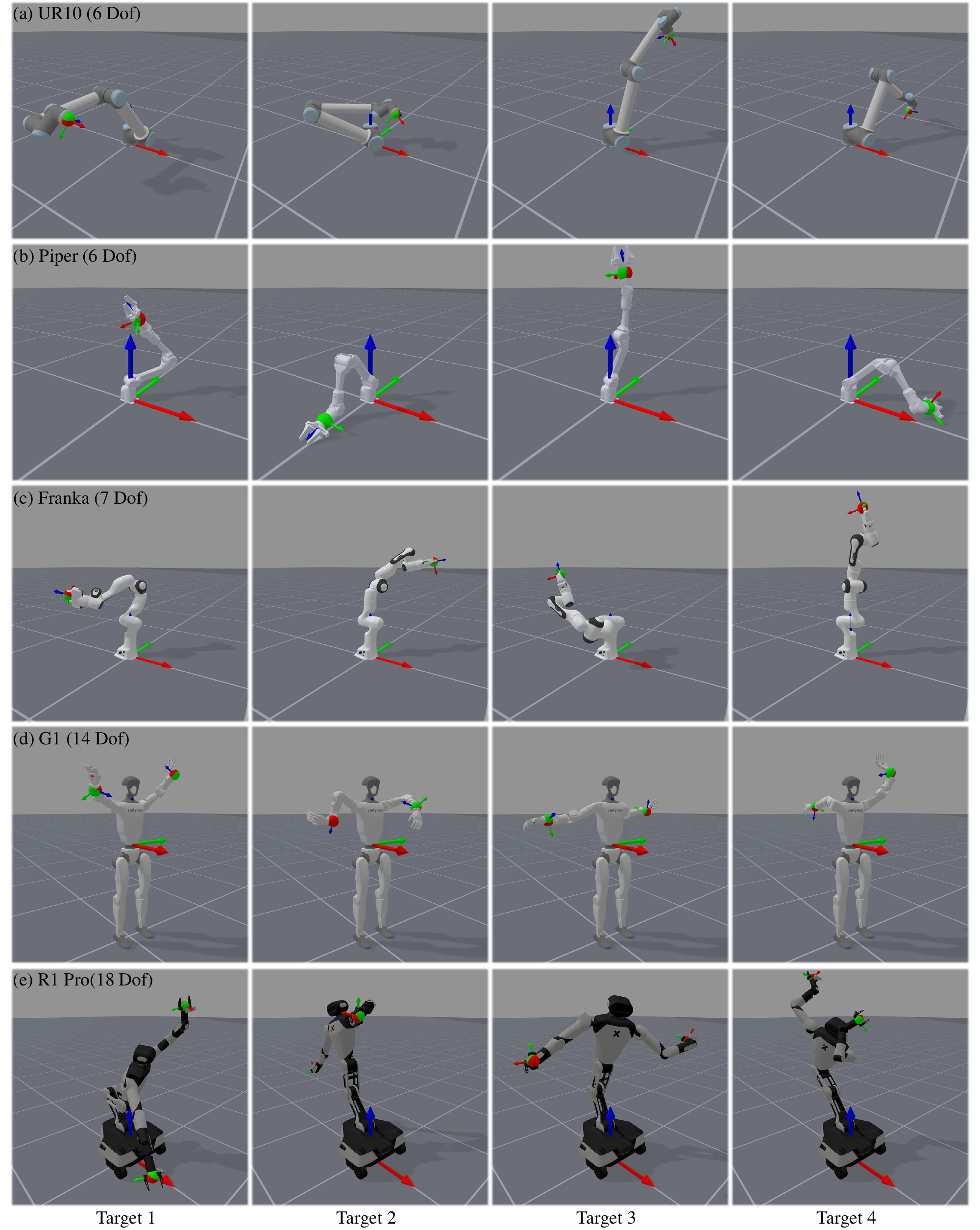}
    \caption{\textbf{Generalization across different robot morphologies and target poses.}
    Each row corresponds to a different robotic platform, while each column represents a different target end-effector pose. The generated results demonstrate that GraphDiff-IK can produce valid inverse kinematics solutions across diverse robot structures and workspace regions while preserving articulated structural consistency.}
    \label{fig:workspace}
\end{figure*}

These results demonstrate that the proposed framework generalizes effectively across diverse robot morphologies and kinematic complexities within a unified graph-based formulation. To further evaluate the generalization capability of the proposed framework across different target poses and workspace regions, we visualize inverse kinematics solutions generated for multiple target end-effector poses across different robot platforms, as shown in Fig.~\ref{fig:workspace}.

Each row in Fig.~\ref{fig:workspace} corresponds to a different robotic platform, while each column represents a different target end-effector pose. The generated results demonstrate that GraphDiff-IK can produce valid inverse kinematics solutions across a wide range of workspace regions and robot structures. For single-arm manipulators, the generated solutions exhibit accurate end-effector alignment across different target poses while maintaining physically plausible articulated configurations. For dual-arm and humanoid systems, the proposed framework additionally preserves coordinated multi-branch motion and torso or waist consistency during inference. These results further demonstrate that the proposed graph diffusion formulation can generalize effectively across diverse robot morphologies and target pose distributions.

\subsection{Multi-Solution Inverse Kinematics Analysis}

\begin{figure*}[!t]
    \centering
    \includegraphics[width=0.99\linewidth]{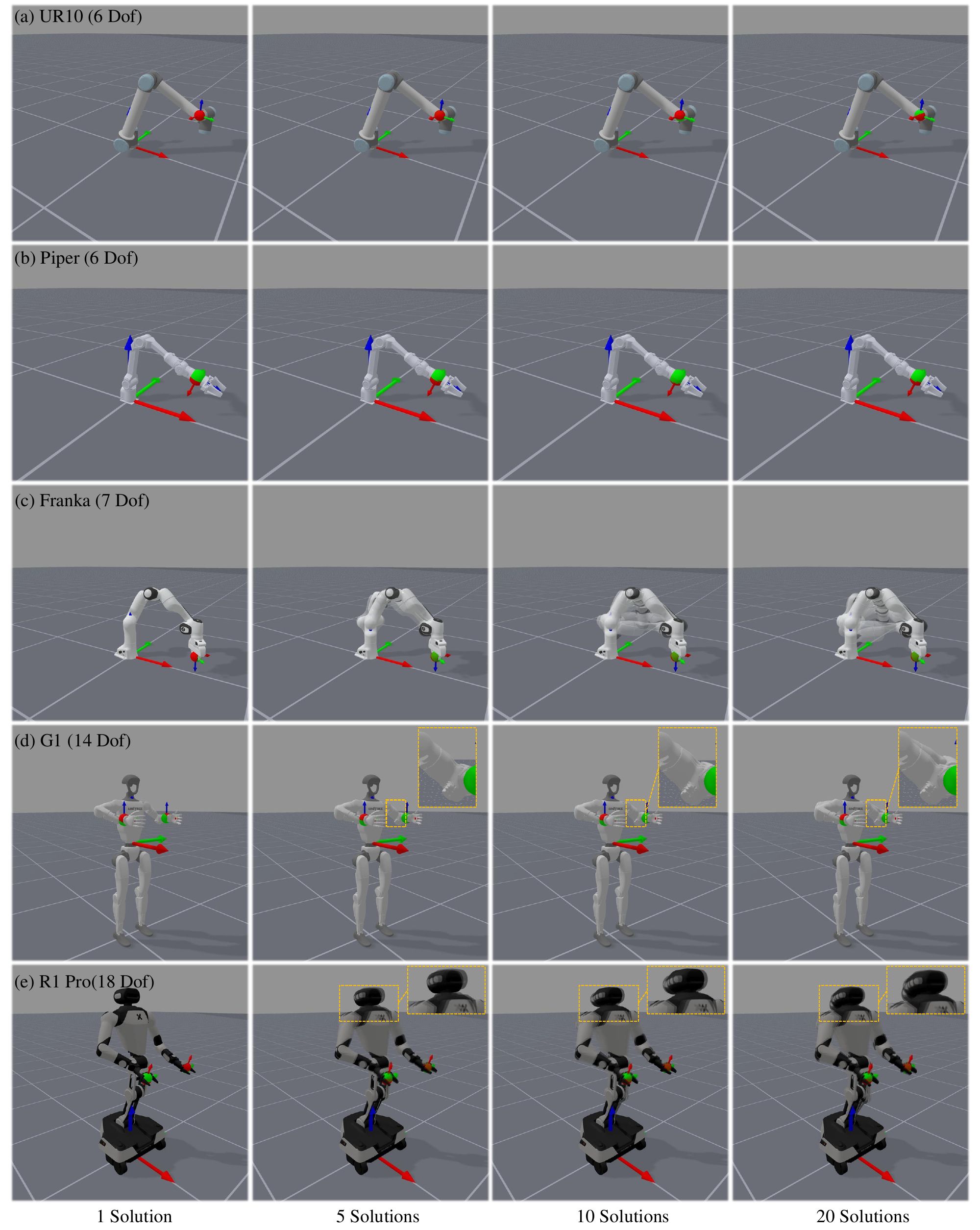}
    \caption{\textbf{Visualization of multiple inverse kinematics solutions generated by GraphDiff-IK.}
    Each row corresponds to a robot platform, while each column shows 1, 5, 10, and 20 generated solutions overlaid for the same target end-effector pose. Non-redundant robots exhibit highly overlapping solutions, whereas redundant and multi-branch systems produce diverse valid articulated configurations for the same target pose.}
    \label{fig:multi_solution}
\end{figure*}

One important characteristic of inverse kinematics is the existence of multiple valid joint configurations that satisfy the same end-effector target pose. This phenomenon is particularly significant in redundant and multi-branch robotic systems, where additional degrees of freedom introduce larger and more flexible solution spaces. To evaluate whether the proposed framework can capture such multi-modal inverse kinematics distributions, we visualize multiple generated solutions for the same target pose across different robot platforms. Fig.~\ref{fig:multi_solution} presents multiple inverse kinematics solutions generated by GraphDiff-IK for the same target pose. Each row corresponds to a different robot platform, while each column visualizes 1, 5, 10, and 20 generated solutions overlaid in the same workspace.

For non-redundant manipulators, including AgileX Piper and Universal Robots UR10, the generated solutions largely overlap even as the number of generated samples increases. This behavior indicates that the inverse kinematics solution space is highly constrained and admits nearly unique valid configurations for a given target pose.

In contrast, redundant and multi-branch robotic systems, including Franka Emika Panda, Unitree G1, and Galaxea R1 Pro, exhibit substantially more diverse solution distributions. As the number of generated samples increases, the proposed framework generates multiple distinct articulated configurations that all satisfy the same end-effector constraint. These results suggest that the proposed graph diffusion formulation naturally captures the multi-modal structure of inverse kinematics solution spaces without requiring explicit multi-solution supervision.

The observed behavior is also consistent with inverse kinematics theory. Non-redundant manipulators typically admit limited feasible solutions for a given target pose, whereas redundant and articulated multi-branch systems possess additional null-space flexibility, allowing multiple valid joint configurations to satisfy the same end-effector constraint. Overall, these results demonstrate that GraphDiff-IK can effectively model the multi-modal nature of inverse kinematics while preserving kinematic feasibility and articulated structure.

\subsection{Interpretability of the Denoising Process}

To further analyze the behavior and interpretability of the proposed graph diffusion framework, we visualize the iterative denoising process during inference. Starting from Gaussian noise in the joint configuration space, the model progressively refines noisy joint states into feasible inverse kinematics solutions through iterative denoising steps.

\begin{figure*}[!t]
    \centering
    \includegraphics[width=0.99\linewidth]{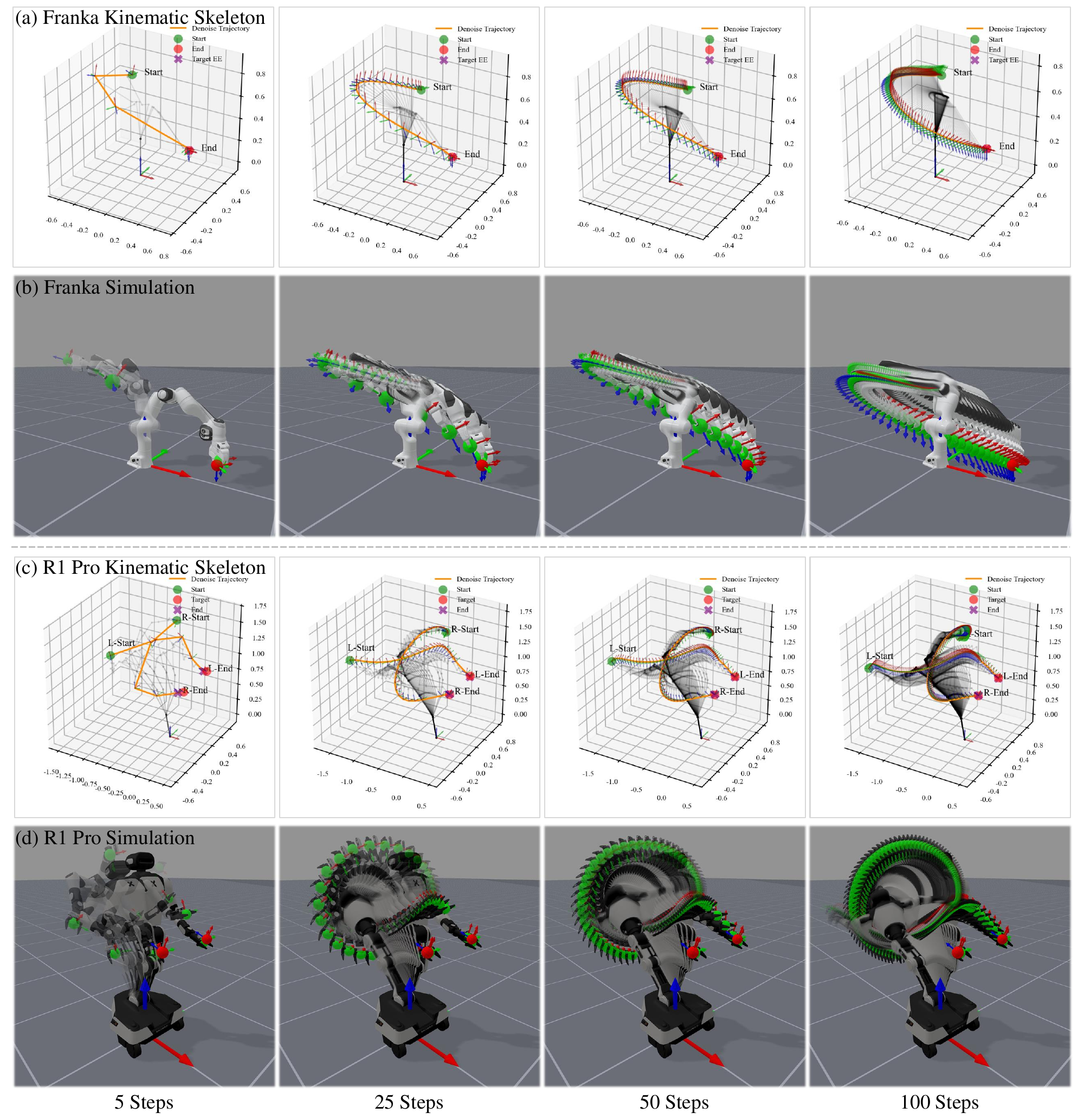}
    \caption{\textbf{Visualization of the iterative denoising process in GraphDiff-IK.}
    Each column corresponds to different denoising steps (5, 25, 50, and 100). The first row for each robot shows the articulated kinematic skeleton reconstructed via forward kinematics, while the second row visualizes overlaid simulation trajectories during diffusion inference.}
    \label{fig:denoising}
\end{figure*}

Fig.~\ref{fig:denoising} visualizes the denoising trajectories generated by GraphDiff-IK for different robot platforms. Each column corresponds to a different denoising step, while each robot is visualized using two rows. The first row shows the articulated kinematic skeleton reconstructed using forward kinematics, illustrating the structural evolution of robot configurations in 3D space. The second row presents overlaid simulation trajectories with transparency, providing a more intuitive visualization of the convergence behavior during inference.

As shown in Fig.~\ref{fig:denoising}, the proposed framework gradually transforms initially noisy and physically implausible joint configurations into structured and kinematically feasible robot poses. During the early denoising stages, robot configurations exhibit substantial structural uncertainty and large deviations from the target end-effector pose. As denoising progresses, the generated configurations progressively converge toward valid articulated structures and accurate end-effector alignment. Interestingly, even with a relatively small number of denoising steps, the generated configurations already approach feasible inverse kinematics solutions. Increasing the number of denoising steps further improves geometric consistency and structural stability, resulting in smoother and more accurate articulated configurations. This behavior suggests that the proposed framework effectively captures the inverse kinematics solution space.

Overall, these visualizations provide intuitive evidence that GraphDiff-IK performs structured iterative refinement during inference rather than directly regressing joint configurations in a single step. The progressive denoising behavior further highlights the interpretability and stability of the proposed diffusion-based inverse kinematics formulation across diverse robot morphologies.

\subsection{Discussion}

Inverse kinematics is inherently a multi-modal problem, particularly in redundant and multi-branch robotic systems, where multiple valid joint configurations may satisfy the same end-effector constraint. Traditional inverse kinematics approaches typically formulate the problem as regression or iterative optimization, often overlooking the underlying structure and diversity of the solution space. In contrast, the proposed GraphDiff-IK formulates inverse kinematics as a conditional structured generation problem over robot kinematic graphs, enabling the model to naturally capture the multi-solution characteristics of inverse kinematics.

Experimental results demonstrate that the proposed framework exhibits behavior consistent with inverse kinematics theory across different robot morphologies. For non-redundant manipulators, the generated solutions converge toward nearly unique configurations, while redundant and multi-branch systems exhibit diverse valid solutions for the same target pose. In addition, the proposed graph-based formulation generalizes effectively across robots with varying degrees of freedom and kinematic topologies, including single-arm manipulators, dual-arm systems, and humanoid robots with torso or waist articulation. These observations suggest that explicitly incorporating robot structure through graph representations plays an important role in modeling articulated kinematic dependencies and improving scalability across complex robotic systems.

Despite the promising results, the proposed framework still has several limitations. First, diffusion-based iterative denoising introduces higher inference cost compared to direct regression methods. Second, the current formulation focuses primarily on kinematic feasibility and does not explicitly consider collision avoidance, dynamic constraints, or temporal motion consistency. Future work will focus on improving inference efficiency and extending the proposed framework toward constraint-aware whole-body motion generation and real-world robotic manipulation tasks.


\section{Conclusion}

In this paper, we proposed GraphDiff-IK, a structure-aware graph diffusion framework for inverse kinematics. By representing robots as kinematic graphs and formulating inverse kinematics as a conditional diffusion-based generation problem, the proposed method explicitly incorporates robot topology into the learning and inference process. Unlike conventional deterministic regression approaches, GraphDiff-IK naturally models the multi-modal nature of inverse kinematics and is capable of generating multiple valid joint configurations for the same target end-effector pose.

Experimental results on a diverse set of robotic platforms demonstrate that the proposed framework achieves accurate and stable inverse kinematics performance across different robot morphologies, including single-arm manipulators, dual-arm systems, and humanoid robots with torso or waist articulation. In addition, qualitative visualizations further show that the proposed method captures meaningful multi-solution inverse kinematics behavior and exhibits interpretable progressive denoising dynamics during inference.

Overall, the proposed graph diffusion formulation provides a unified and scalable framework for inverse kinematics across diverse articulated robotic systems. Future work will focus on improving inference efficiency and extending the framework toward collision-aware motion generation, temporal trajectory modeling, and real-world whole-body robotic manipulation tasks.

\bibliographystyle{IEEEtranBST/IEEEtran.bst}
\bibliography{IEEEtranBST/IEEEabrv, bib}

\end{document}